\documentclass[conference]{IEEEtran}
\IEEEoverridecommandlockouts
\usepackage{cite}
\usepackage{amsmath,amssymb,amsfonts}
\usepackage{algorithmic}
\usepackage{graphicx}
\usepackage{textcomp}
\usepackage{xcolor}
\def\BibTeX{{\rm B\kern-.05em{\sc i\kern-.025em b}\kern-.08em
    T\kern-.1667em\lower.7ex\hbox{E}\kern-.125emX}}
\begin{document}

% 기존 자기가 하던 연구에서 minor한 results 내용으로 작성
% 저자에 교수님은 전부 다 제외
% 저자는 최대 3명까지만 허용
% 분량은 4장 꽉
% 10월 18일 금요일 오후 2시까지 개인톡으로 PDF VERSION 송부 (REFERENCE 파트에 저희 연구실 논문 하이라이트 쳐서)
% 홈페이지 링크 (https://brain.korea.ac.kr/bci2025/index.php)
% 본문에 단위 있을 경우, 단위 앞에 space 하나씩

\title{Imagined Speech State Classification for Robust Brain-Computer Interface
%{\footnotesize \textsuperscript{*}Note: Sub-titles are not captured for https://ieeexplore.ieee.org  and should not be used}
\thanks{This work was partly supported by Institute of Information \& Communications Technology Planning \& Evaluation (IITP) grant funded by the Korea government (MSIT) (No. RS--2021--II--212068, Artificial Intelligence Innovation Hub, No. RS--2024--00336673, AI Technology for Interactive Communication of Language Impaired Individuals, and No. RS--2019--II190079, Artificial Intelligence Graduate School Program (Korea University)).}}
% Dept. of Brain and Cognitive Engineering
% Dept. of Artificial Intelligence
% Harnessing Deep Learning for Imagined Speech Recognition in BCI Systems
% Deep Learning-Driven Imagined Speech Recognition for Brain-Computer Interfaces
% Deep Learning Approaches for Imagined Speech Interpretation in Brain-Computer Interfaces

\author{
\IEEEauthorblockN{Byung-Kwan Ko}
\IEEEauthorblockA{\textit{Dept. of Artificial Intelligence} \\
\textit{Korea University}\\
Seoul, Republic of Korea \\
leaderbk525@korea.ac.kr} \\
\and

\IEEEauthorblockN{Jun-Young Kim}
\IEEEauthorblockA{\textit{Dept. of Artificial Intelligence} \\
\textit{Korea University}\\
Seoul, Republic of Korea \\
j\_y\_kim@korea.ac.kr}
\and

\IEEEauthorblockN{Seo-Hyun Lee}
\IEEEauthorblockA{\textit{Dept. of Brain and Cognitive Engineering} \\
\textit{Korea University} \\
Seoul, Republic of Korea \\
seohyunlee@korea.ac.kr}\\
}

\maketitle

\begin{abstract}
This study examines the effectiveness of traditional machine learning classifiers versus deep learning models for detecting the imagined speech using electroencephalogram data. Specifically, we evaluated conventional machine learning techniques such as CSP-SVM and LDA-SVM classifiers alongside deep learning architectures such as EEGNet, ShallowConvNet, and DeepConvNet. Machine learning classifiers exhibited significantly lower precision and recall, indicating limited feature extraction capabilities and poor generalization between imagined speech and idle states. In contrast, deep learning models, particularly EEGNet, achieved the highest accuracy of 0.7080 and an F1 score of 0.6718, demonstrating their enhanced ability in automatic feature extraction and representation learning, essential for capturing complex neurophysiological patterns. These findings highlight the limitations of conventional machine learning approaches in brain-computer interface (BCI) applications and advocate for adopting deep learning methodologies to achieve more precise and reliable classification of detecting imagined speech. This foundational research contributes to the development of imagined speech-based BCI systems.
\end{abstract}


\section{INTRODUCTION}
Brain-computer interfaces (BCIs) have emerged as a transformative technology, offering unprecedented avenues for communication and control by translating neural activity into actionable commands\cite{diez2011asynchronous, chaudhary2016brain}. Among the various paradigms within BCI research, imagined speech-based systems represent a particularly promising frontier\cite{9268982}. Unlike traditional methods that rely on external stimuli\cite{lopes2019online}, imagined speech allows users to generate linguistic representations internally, enabling more natural and intuitive interactions\cite{lee2023AAAI}. This capability has significant potential to improve communication for people with severe motor disabilities and to expand the applications of BCIs in broader, real-world contexts.

However, the intricate nature of speech-related brain mechanisms presents substantial challenges for the development of BCI. Speech production and perception involve complex coordinated neural processes in multiple brain regions, making the accurate decoding of imagined speech signals from electroencephalogram (EEG) data extremely demanding\cite{panwar2020modeling}. The temporal and spatial dynamics of neural activity associated with speech are subtle and variable, necessitating sophisticated analytical techniques to distinguish meaningful patterns from noise\cite{wang2013analysis}. As a result, obtaining high classification accuracy in imagined speech tasks remains a significant challenge, hindering the practical implementation of such systems.

Endogenous BCI research has traditionally relied on machine learning (ML) techniques, which, despite their contributions, have certain limitations when applied to the complex nature of EEG signals. Conventional ML algorithms, such as support vector machines (SVM) and linear discriminant analysis (LDA), often rely on manually engineered features and simplified models that struggle to capture the intricate temporal and spatial dynamics inherent in neural data. Imagined speech decoding, in particular, involves highly nuanced and subtle signal variations, making it difficult for conventional ML models to extract meaningful patterns without significant data preprocessing or tailored feature engineering. This dependence on feature crafting can lead to suboptimal generalization, especially across subjects or tasks, ultimately limiting the responsiveness and adaptability of ML-driven BCI systems for imagined speech.

\begin{figure*}[t]
\centering
    \includegraphics[width=\textwidth]{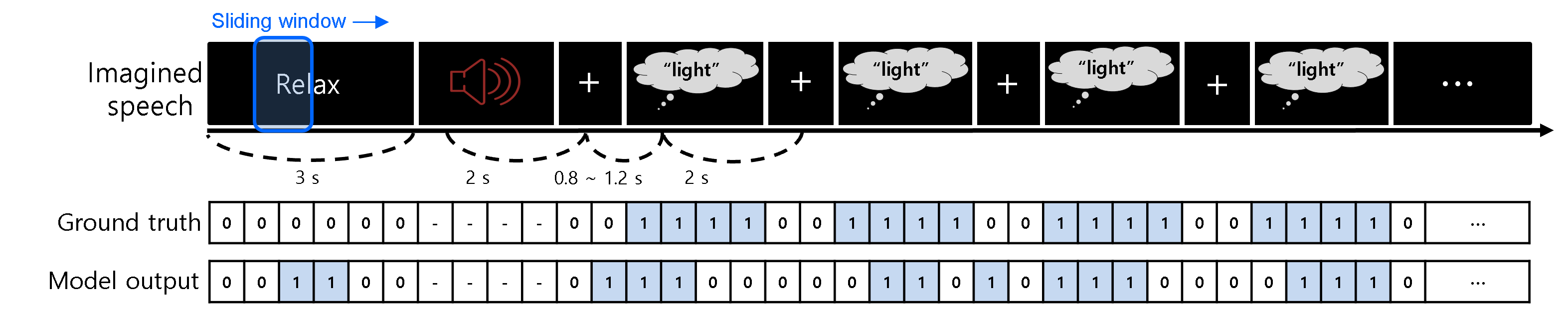}
    \caption{Overall scheme of proposed method.}
    \label{figure1}    
    \vspace{-0.3cm}
\end{figure*}

In contrast, deep learning (DL) offers transformative potential by automatically learning complex features directly from raw EEG data\cite{9606552}, thus reducing the need for extensive manual preprocessing. Deep neural networks, such as convolutional neural networks and recurrent neural networks, excel in modeling temporal and spatial dependencies within data, a capability that is especially beneficial for the intricate nature of imagined speech\cite{han2020enhanced}. Using the layers within these networks, DL approaches can extract hierarchical and high-dimensional representations of neural activity, capturing subtle signal differences that ML models might overlook. Furthermore, DL techniques have demonstrated impressive robustness in various domains, showing promise for improved generalization in subject-independent BCI applications\cite{jeong2019classification, kwon2019subject}. As researchers integrate DL into imagined speech BCIs, they open new avenues for building more accurate, adaptive, and efficient interfaces, advancing the goal of real-world usability and seamless interaction. This transition marks a significant step forward, addressing the limitations of traditional ML and fully capitalizing on the rich information embedded in neural signals associated with imagined speech.

In this study, we aim to advance imagined speech-based BCI research by leveraging DL to decode neural signals associated with imagined speech more accurately and robustly\cite{watanabe2020synchronization}. Departing from traditional ML approaches, we investigate the application of DL architectures to improve the accuracy and robustness of imagined speech decoding from EEG signals. By leveraging DL's advanced capacity for extracting complex, high-dimensional features directly from neural data, our research seeks to create a more effective framework for imagined speech BCIs. Moreover, we underscore the limited research on DL in this specific domain, highlighting the novelty and potential impact of our approach. Our findings underscore the effectiveness of DL in enhancing performance, marking a step forward in the development of imagined speech-based BCIs for future applications.
\section{MATERIALS AND METHODS}

\subsection{Data Description}
The dataset employed in this study was obtained from the research conducted by Lee et al.~\cite{lee2020neural, jeong20222020}. For the preliminary analysis, EEG data from six subjects were utilized. EEG signals were recorded using a 64-channel actiCAP equipped with active Ag-AgCl electrodes, ensuring high-quality signal acquisition and minimizing artifacts. The FCz and FPz electrodes were designated as the reference and ground electrodes, respectively. Participants were instructed to perform tasks involving both imagined speech and overt speech, encompassing twelve distinct words and one idle state. These words were meticulously selected to enhance the potential applicability of the BCI communication system in real-world scenarios, as they are commonly used by patients. EEG signals were amplified using a BrainProduct GmbH (Germany) EEG signal amplifier, with a sampling rate of 1,000 Hz. During overt speech sessions, participants' spoken voices were simultaneously recorded via a microphone with a sampling rate of 16,000 Hz to provide ground truth data. This dual recording approach ensured the accurate alignment of neural signals with corresponding speech activities.\\

\begin{table}[t]
    \centering
    \renewcommand{\arraystretch}{1.5} % Increase row height for better readability
    \setlength{\tabcolsep}{10pt} % Add column spacing for clarity
    \caption{Average Number of Data Samples per Class for Training and Testing.}
    \begin{tabular}{c|c|c}
        \hline
        Class     &     Training Samples     &     Testing Samples     \\ \hline
        Imagined Speech       & 53,293               & 13,325     \\ \hline
        Idle State            & 25,580               & 10,492      \\ \hline
    \end{tabular}
    \label{table1}
\end{table}

\subsection{Experiment Method}
The methodology employed both traditional ML techniques and advanced DL architectures to achieve robust classification between imagined speech and idle states\cite{song2017novel, 10411116}. As to represent ML classifiers, common spatial patterns (CSP) combined with SVM\cite{1004136} and linear discriminant analysis were implemented, leveraging their proven effectiveness in EEG signal classification\cite{9133061}. The study then examined deep learning models, such as EEGNet \cite{lawhern2018eegnet}, ShallowConvNet \cite{schirrmeister2017deep}, and DeepConvNet \cite{schirrmeister2017deep}, which are well-regarded for their ability to automatically extract complex spatial and temporal features from raw EEG data \cite{9175874}.

As shown in Fig.~\ref{figure1}, each 2-second trial was divided into overlapping 500 ms windows with a 50 ms shift, generating a substantial number of samples for model training. Labeling was performed based on specific task intervals: windows corresponding to the imagined speech periods of the 12 target words were labeled as speech state (1), while those within rest periods or cross-mark (+) display intervals were assigned as idle state (0). This cue-based approach enabled the models to effectively learn the temporal distinctions between speech-related neural activity and idle states. By creating a large, labeled dataset from overlapping windows, this method provided an adequately representative distribution of classes, facilitating more robust training on the temporal dynamics of imagined speech versus rest states.

\begin{table}[t]
    \centering    
    \renewcommand{\arraystretch}{1.5} % Increase the row height
    \setlength{\tabcolsep}{10pt} % Add column spacing
    \caption{Comparison of Classification Performance with Different Methods.}
    \resizebox{\columnwidth}{!}{%
    \begin{tabular}{c|cccc}
    \hline
    \textbf{Method}          & \textbf{Accuracy} & \textbf{Precision} & \textbf{Recall} & \textbf{F1-Score} \\ \hline
    CSP-SVM   
    & 0.5890           & 0.5256             & 0.5391             & 0.5323  \\ \hline
    CSP-LDA   
    & 0.5986           & 0.5367             & 0.5186             & 0.5275  \\ \hline
    ShallowConvNet\cite{schirrmeister2017deep}
    & 0.6845           & 0.6531             & 0.6467             & 0.6499  \\ \hline
    DeepConvNet\cite{schirrmeister2017deep}
    & 0.6581           & 0.6254             & 0.6172             & 0.6231  \\ \hline
    \textbf{EEGNet}\cite{lawhern2018eegnet}
    & \textbf{0.7080}           & \textbf{0.6757}             & \textbf{0.6679}   & \textbf{0.6718}  \\
    \hline
    \end{tabular}%
    }
    \label{table2}
\end{table}

Segmented and labeled data were divided into training and testing sets, with 80\% allocated for training and 20\% for testing, using a fixed random seed of 2,024 to ensure reproducibility. Variations in experimental conditions during data collection caused slight imbalances in class ratios, as shown in Table~\ref{table1}. To address these discrepancies, class weights were applied during the training process, enhancing the model's ability to effectively handle imbalanced classes. Traditional ML classifiers used manually extracted CSP features\cite{895946, devlaminck2011multisubject}, while DL models automatically learned the relevant features from raw EEG signals. Each model was then tested on the evaluation set to distinguish between imagined speech and idle states. Performance was assessed using metrics such as accuracy, precision, recall, and F1-score, offering a thorough evaluation of each model’s effectiveness \cite{9007451}.

\section{RESULTS AND DISCUSSION}

The present study first examined the distribution of data samples used for training, noting a imbalance between imagined speech states and idle states. Despite this imbalance, the model demonstrated its capability to classify imagined speech effectively. The performance results further indicated that traditional ML classifiers exhibited lower performance, whereas DL-based models achieved comparatively higher performance. 

The performance results revealed that traditional machine learning classifiers, such as CSP-SVM and LDA-SVM, struggled in accurately classifying imagined speech states, achieving lower precision values of 0.5256 and 0.5367, respectively. Despite the balanced distribution of speech and idle states in the test set, these classifiers showed limited ability to accurately predict either state, suggesting that the training imbalance, with speech samples outnumbering idle samples by approximately two to one, may have affected the classifiers' capacity to generalize well to a balanced distribution. This imbalance likely led to insufficient feature learning for both speech and idle states, resulting in poor generalization on the test set. In contrast, DL models—specifically EEGNet, ShallowConvNet, and DeepConvNet—demonstrated substantial improvements in classification accuracy and balanced state detection. EEGNet, in particular, achieved the best results with an accuracy of 0.7080, a precision of 0.6757, and an F1-score of 0.6718, underscoring its capacity to automatically learn and differentiate between complex EEG features associated with both speech and idle states. These findings suggest that DL models, with their advanced feature extraction capabilities, are more effective at handling imbalances during training and generalizing across class distributions, making them better suited for nuanced EEG-based classification tasks than traditional ML approaches. 

\begin{figure}[ht]
    \centering
    \includegraphics[width=0.9\columnwidth]{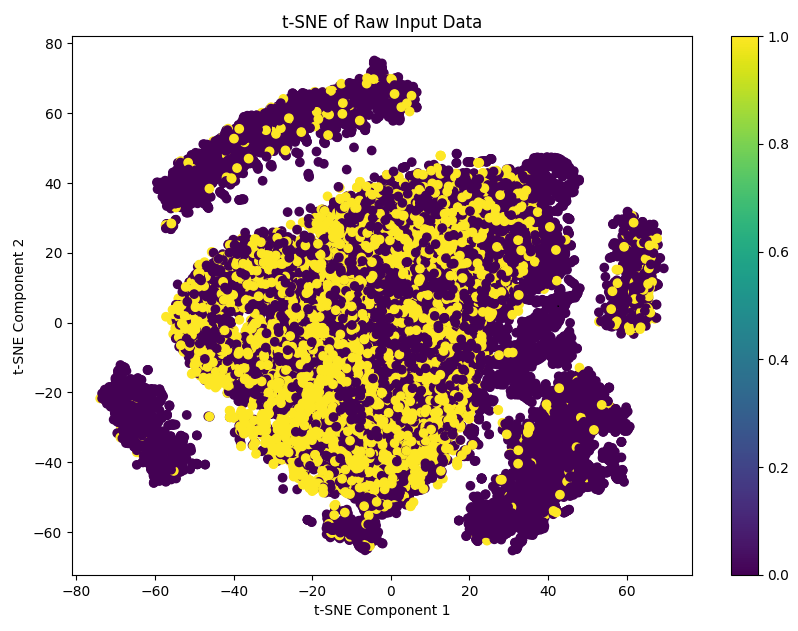}
    \label{figure2}
    \vspace{-0.3cm} % Adjust vertical space as needed
\end{figure}

% Second Figure
\begin{figure}[ht]
    \centering
    \includegraphics[width=0.9\columnwidth]{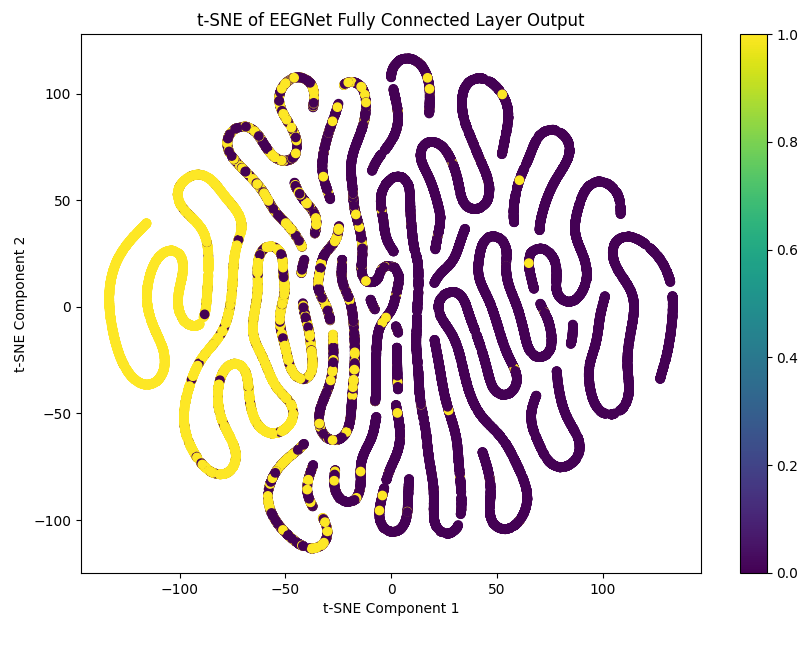}
    \caption{T-distributed stochastic neighbor embedding (t-SNE) visualization of the input data and fully connected layer output of subject 6 via EEGNet.}
    \label{figure3}
    \vspace{-0.3cm} % Adjust vertical space as needed
\end{figure}

These findings underscore the limitations of conventional ML approaches in handling complex neurophysiological data and emphasize the advanced feature learning and representation capabilities inherent to DL architectures\cite{HAN2020324}, facilitating more accurate and reliable imagined speech classification.

Fig.~\ref{figure3} presents the t-SNE visualization of data samples during the training process of EEGNet. Initially, the input data components for the two labels were densely overlapping, making them difficult to distinguish. However, as the model trained, the visualization of the final fully connected layer revealed that the data points became partially separable. This separation reflects successful feature learning, indicating that the model effectively captured distinct spatial and temporal patterns from the raw EEG signals linked to imagined speech.

In the cue-based labeling process, we acknowledged that each 2-second trial window does not represent a continuous imagined speech state. Only certain segments within this window likely capture the true imagined speech phase where subjects actively internalize the target word. Given the inherent nature of imagined speech, pinpointing the exact onset and offset of this internal speech remains challenging. However, further aligning ground truth labels more precisely with the actual imagined speech intervals within each trial could lead to improved model performance by providing a more accurate representation of speech-related neural dynamics\cite{proix2022imagined}. This refined labeling would also help balance the sample counts between imagined speech and idle states, creating a more balanced dataset that is conducive to effective model training.

Additionally, this approach leverages the potential benefit of using overt speech neural mechanisms as a reference to detect shared patterns within imagined speech, given their overlapping characteristics\cite{soroush2023nested}. Future work could focus on enhancing this alignment technique, which may bring substantial advancements in imagined speech-based BCI performance. By drawing on the common neural substrates between overt and imagined speech, this refined labeling method promises to improve model accuracy and robustness, especially in distinguishing imagined speech from idle states.

\section{CONCLUSION}
This study explored the comparative effectiveness of traditional ML classifiers and DL models in classifying imagined speech using EEG data within a BCI framework. The findings revealed that traditional classifiers struggled with low precision and recall, underscoring their limitations in feature extraction and in handling the complex neurophysiological patterns associated with imagined speech. In contrast, DL models, particularly EEGNet, demonstrated a superior ability to automatically extract and learn intricate features from EEG signals, resulting in more accurate and reliable classification. These advancements highlight the potential of DL methodologies to address the shortcomings of conventional ML approaches in BCI applications and pave the way for more effective imagined speech-based BCI systems. Future work may focus on refining DL architectures to further enhance classification accuracy, while exploring methods to better align imagined speech intervals for improved labeling and model training. Additionally, validating these advanced models in real-world environments will be essential to advancing the practical implementation of BCI systems, ultimately bringing imagined speech-based BCIs closer to robust, real-world applications.

%\section*{Acknowledgment}

\bibliography{References}

\bibliographystyle{IEEEtran}

%- 저자 5명까지는 전부 다 적기
%- 저자 6명 이상부터는 1저자 + et al. 처리하기
%- journal 명은 abbreviation iso4로 작성하기
%- conference 명은 줄일 수 있는 부분 최대한 줄이고 뒤에 약자로 뭐라고 표시하는지 표기하기
% 우리 연구실 논문 레퍼런스로 넣을 때 최대한 연달아 안 나오게끔 배치하기
% 레퍼런스 타이틀이 "~~~: ~~~"의 구성을 띌 경우, ":" 다음에 등장하는 단어 맨 앞 대문자

\end{document}